\documentclass{article}

\usepackage{arxiv}

\usepackage[utf8]{inputenc} 
\usepackage[T1]{fontenc}    
\usepackage{hyperref}       
\usepackage{url}            
\usepackage{booktabs}       
\usepackage{amsfonts}       
\usepackage{nicefrac}       
\usepackage{microtype}      
\usepackage{lipsum}
\usepackage{graphicx}
\graphicspath{ {./images/} }
\usepackage{amsmath}
\usepackage{multirow}

\title{Hybrid Interpretable Deep Learning Framework for Skin Cancer Diagnosis: Integrating Radial Basis Function Networks with Explainable AI}

\author{
 Mirza Ahsan Ullah \\
  Department of Computer Science\\
  COMSATS University Islamabad\\
  Islamabad, Paksitan\\
  \texttt{engr.ahsanmirza@gmail.com} \\
   \And
 Tehseen Zia\\
  Department of Computer Science\\
  COMSATS University Islamabad\\
  Islamabad, Paksitan\\
  \texttt{tehseen.zia@comsats.edu.pk} \\
}

\begin{document}
\maketitle
\begin{abstract}
Skin cancer is one of the most prevalent and potentially life-threatening diseases worldwide, necessitating early and accurate diagnosis to improve patient outcomes. Conventional diagnostic methods, reliant on clinical expertise and histopathological analysis, are often time-intensive, subjective, and prone to variability. To address these limitations, we propose a novel hybrid deep learning framework that integrates convolutional neural networks (CNNs) with Radial Basis Function (RBF) Networks to achieve high classification accuracy and enhanced interpretability. The motivation for incorporating RBF Networks lies in their intrinsic interpretability and localized response to input features, which make them well-suited for tasks requiring transparency and fine-grained decision-making. Unlike traditional deep learning models that rely on global feature representations, RBF Networks allow for mapping segments of images to chosen prototypes, exploiting salient features within a single image. This enables clinicians to trace predictions to specific, interpretable patterns. The framework incorporates segmentation-based feature extraction, active learning for prototype selection, and K-Medoids clustering to focus on these salient features. Evaluations on the ISIC 2016 and ISIC 2017 datasets demonstrate the model's effectiveness, achieving classification accuracies of 83.02\% and 72.15\% using ResNet50, respectively, and outperforming VGG16-based configurations. By generating interpretable explanations for predictions, the framework aligns with clinical workflows, bridging the gap between predictive performance and trustworthiness. This study highlights the potential of hybrid models to deliver actionable insights, advancing the development of reliable AI-assisted diagnostic tools for high-stakes medical applications.
\end{abstract}



\section{Introduction}
Skin cancer is among the most prevalent and potentially fatal diseases worldwide, with melanoma and non-melanoma types contributing significantly to the global health burden. Early detection and precise diagnosis are pivotal for improving patient outcomes and reducing mortality rates. Conventional diagnostic approaches heavily rely on clinical expertise and histopathological analysis, which, despite their utility, are often subjective, labor-intensive, and prone to inter-observer variability. This necessitates the development of automated and reliable systems to assist in skin cancer diagnosis. In this context, machine learning (ML) algorithms, particularly deep learning models, have emerged as promising tools for leveraging image-based diagnostic data to enhance accuracy and efficiency \cite{1}.

\par
Deep learning, a subset of ML, has achieved state-of-the-art results in numerous image classification and segmentation tasks, including medical applications such as skin lesion analysis. However, as these models become more complex, their decision-making processes become increasingly opaque, widely known as the "Black Box" problem. This opacity limits the usability of deep learning models in sensitive domains like healthcare, where interpretability and trustworthiness are critical. For diagnostic systems, clinicians must rely on the predictions and understand the rationale behind them to align the model's output with clinical knowledge and standards of care. As a result, regulatory guidelines such as those outlined by the European Union's General Data Protection Regulation (GDPR) have emphasized the need for transparency and explainability in artificial intelligence (AI) systems deployed in high-stakes applications \cite{2,3,4}.

\par
To address the interpretability challenge, numerous post-hoc explainability methods, such as SHAP (Shapley Additive explanations), LIME (Local Interpretable Model-agnostic Explanations), and saliency-based techniques like Grad-CAM (Gradient-weighted Class Activation Mapping), have been proposed \cite{5,6,7}. These methods provide information on the model's behavior by generating visual or numerical explanations of the predictions. However, post hoc techniques often approximate the reasoning process of a black-box model, introducing assumptions and simplifications that may not faithfully replicate the actual computations of the network \cite{8,9}. This behavior of post-hoc techniques can lead to inconsistent or unreliable explanations, mainly when different methods yield disparate interpretations for the same input \cite{10}. Moreover, interpreting the results of post-hoc methods requires additional expertise, further complicating their adoption in clinical workflows. As discussed in \cite{11}, inherently interpretable models, which are transparent by design, offer a more robust alternative to mitigate the limitations of post-hoc explainability.

\par
To address these challenges, we propose a novel hybrid deep learning framework integrating a Radial Basis Function Neural Network (RBF-NN) with convolutional neural networks (CNNs). The RBF-NN is intrinsically interpretable due to its localized response to input features, making it well-suited for providing transparent explanations. The training process incorporates a deep active learning paradigm, where critical concepts extracted by experts from the dataset are identified as centroids for the RBF neurons using K-medoids clustering. These centroids serve as the building blocks for classification and explanation tasks, ensuring that the model's decisions are faithful to the data and aligned with human reasoning. Unlike traditional models that provide global predictions, our approach averages decisions across localized patches within an image, enabling a fine-grained analysis that mirrors the diagnostic process used by clinicians.

\par
The proposed framework is validated in benchmark datasets, including ISIC 2016 \cite{gutman2016skin} and ISIC 2017 \cite{codella2018skin}, achieving comparable classification precision and interpretability performance. Using pre-trained CNN embeddings, our approach bridges the gap between representation learning and interpretable modeling, offering a scalable solution for applications with limited training data, as is often the case in medical imaging.

This paper makes the following contributions to the field of medical diagnostics and interpretable deep learning:

\begin{enumerate}
\item Proposes a hybrid deep learning framework integrating convolutional neural networks (CNNs) and Radial Basis Function (RBF) Networks to enhance accuracy and interpretability in skin cancer diagnosis.
\item This research paper introduces a prototype-based decision-making process that maps image segments to selected prototypes, exploiting salient features within individual segments. This approach enhances interpretability and accuracy by prioritizing critical details, significantly contributing to fine-grained image analysis tasks.

\item Demonstrates scalability and adaptability of the framework for broader applications in medical imaging, validated through performance on benchmark datasets (ISIC 2016 and ISIC 2017).
\end{enumerate}

\par
This paper is organized as follows: Section \ref{relatedwork} reviews related work, focusing on explainability challenges in deep learning and current approaches to medical diagnosis. Section \ref{proposedmodel} describes the architecture of the proposed hybrid model, detailing its components, training methodology, and interpretability mechanisms. This section also outlines this study's experimental setup, datasets, and evaluation metrics. Section \ref{results} presents and discusses the results, including a comparison with related non-interpretable and interpretable models. Finally, section \ref{conclusion} concludes the paper, summarizing contributions and identifying directions for future research.

\section{Related Work}\label{relatedwork}
Interpretability has emerged as a pivotal area of research within the machine learning community, particularly in critical domains such as healthcare, where transparent explanations are essential for fostering trust among health personnel. Deep learning models are currently state of the art for numerous healthcare-related tasks, including disease diagnostics and classification \cite{bolhasani2021deep,esteva2019guide,pandey2019recent,kaul2022deep,chakraborty2024machine,maqsood2023multiclass,shafqat2023leveraging}. However, their black-box nature poses significant challenges to their adoption in such sensitive fields. Consequently, the growing demand for interpretable solutions has driven the development of various methods to elucidate model decisions. Broadly, these techniques are categorized as (i) post-hoc methods and (ii) ante-hoc methods.

\subsection{Post-hoc Methods}
Post-hoc methods aim to interpret models after training, often employing visualization techniques to provide local explanations for individual predictions. For instance, methods like Class Activation Mapping (CAM) \cite{zhou2016learning}, Grad-CAM \cite{chattopadhay2018grad}, and Opti-CAM \cite{zhang2024opti} highlight input regions that Deep Neural Networks (DNNs) focus on during predictions. These techniques have been widely applied in interpreting deep learning models. For example, Yang et al. \cite{1} leveraged ResNet-50 \cite{he2016deep} with CAM to classify skin diseases in the ISIC 2017 dataset, achieving an accuracy of 83.00\%. Similarly, Young et al. \cite{2} utilized Grad-CAM and Kernel SHAP for the HAM10000 dataset, attaining an accuracy of 85\%. Zunair et al. \cite{zunair2020melanoma} used VGG-16 \cite{simonyan2014very} with CAM to classify diseases in the ISIC 2016 dataset, achieving a sensitivity of 91.76\% and an accuracy of 81.18\%.

\par
While effective in some contexts, post-hoc methods face notable limitations, especially in healthcare. Their explanations are generated independently of the model, making it difficult to ascertain whether inaccuracies stem from the explainer or the model itself \cite{achtibat2023attribution}. Moreover, these methods require users to trust approximations that might be systematically biased or inconsistent \cite{r-5, r-6}. In domains like pathology, where image features often require multi-faceted classification (e.g., tumor type, stage, and mutations), single-feature-focused techniques like CAM or Grad-CAM fall short. Accurate diagnosis necessitates a comprehensive understanding of diverse tissue characteristics, highlighting the need for advanced interpretability techniques \cite{r-6-18}.

\subsection{Ante-hoc Methods}
Acknowledging these limitations, ante-hoc methods have garnered attention for their ability to learn to explain and predict jointly. Rudin \cite{r-7} and Lipton \cite{r-8} champion inherently interpretable models over post-hoc explanations. Classical models such as decision trees \cite{quinlan1986induction,rokach2005decision}, linear models \cite{khashei2012novel}, and rule-based systems \cite{van2017fifty,glaab2012using} offer straightforward interpretability due to their inherent simplicity. However, these methods often lack scalability and performance when tackling complex medical datasets requiring fine-grain classifications.

\subsection{Prototype-Based Deep Learning Models}
Prototype-based deep learning models represent a promising approach by combining explanation and prediction. Radial Basis Function (RBF) networks \cite{buhmann2000radial} provide a balance between interpretability and functionality. Their localized structure allows for some level of analysis, especially in fields like image processing and robotics \cite{majdisova2017radial}. However, while converging faster than traditional models, their accuracy on complex datasets remains limited.
\par
Several deep-learning approaches have been developed to learn the salient features as prototypes, thus increasing the efficacy of these models while providing local explanations in a form that provides insights into the model inference mechanism. Among such models, a prominent is ProtoPNet \cite{chen2019looks}, which utilizes learned prototypes to map latent features to human-understandable concepts. Recent enhancements, such as ProtoPShare \cite{rymarczyk2021protopshare}, improve efficiency by merging semantically similar prototypes. These advancements provide both accurate predictions and a deeper understanding of class relationships. Despite their potential, current prototype-based methods often rely on single-feature analysis, which limits their ability to address multi-concept datasets. For instance, in medical imaging, where various features like blue veins or other critical markers must be considered collectively, single-concept comparisons fail to offer holistic insights. Recent innovations like ProtoConcepts \cite{ma2023looks} tackle this issue by visualizing prototypes as multi-patch concepts, enabling nuanced interpretations across different features. However, challenges such as limited data availability and the need for concept labeling remain significant, especially in domains like healthcare.

\par
The demand for interpretable models that perform comparably to deep learning systems while requiring minimal data persists. Integrating human-like reasoning into models can further enhance their interpretability and usability. For example, Zia et al. \cite{zia2022softnet} proposed SoftNet, a two-stream network inspired by dual-processing human cognition theories. Its System I employs a shallow CNN for generalization, while System II incorporates a concept memory network for high-level explanations. Building on this, Ullah et al. \cite{ullah2024inherently} introduced CA-SoftNet, which improved accuracy by sharing visually similar concepts across classes. SoftNet and CA-SoftNet represent significant progress, offering human-understandable and faithful explanations. However, parts of these models remain opaque due to their reliance on deep learning subcomponents for generalization. Addressing this "black-box" aspect is critical for future research to achieve models that are accurate, fully interpretable, and reliable for high-stakes applications like healthcare.

\section{Proposed architecture}\label{proposedmodel}
We are proposing a hybrid interpretable deep learning model that integrates representation learning and interpretable machine learning techniques to address the challenges of skin cancer diagnosis. First, each input image is partitioned into segments; later, these segments are transformed into high-dimensional embeddings by leveraging a pre-trained convolutional neural network (CNN), e.g., VGG-16, ResNet-50, etc. These embeddings (features) represent meaningful features of skin lesions, capturing essential patterns and details.

\par
To enhance interpretability and classification performance, users select prominent class-wise features using active learning. Later, clustering is applied using K-medoids to select prominent features as centroids. These features serve as prototypes of the proposed Radial basis function model. The prototypes summarize key patterns in the data, providing a basis for understanding how the model categorizes different lesion types. An RBF (Radial Basis Function) layer further processes the embeddings by computing the similarity of each input embedding to the cluster prototypes. The RBF layer transforms these similarities into a decision-making space where classification occurs. The model's interpretability comes from the ability to analyze cluster memberships and RBF responses, highlighting the relationship between the input data and learned prototypes. This step helps identify the underlying reasoning mechanism by making it transparent to end users. The explanations are generated through the exact mechanism within the model, thus helping to create the user's trust in the proposed model. A graphical description of the model is shown in figure \ref{fig:proposedModel}, given below. A detailed description of the proposed architecture and its components is provided in the following subsections.

\begin{figure}[!ht]
    \centering
    \includegraphics[width=\textwidth]{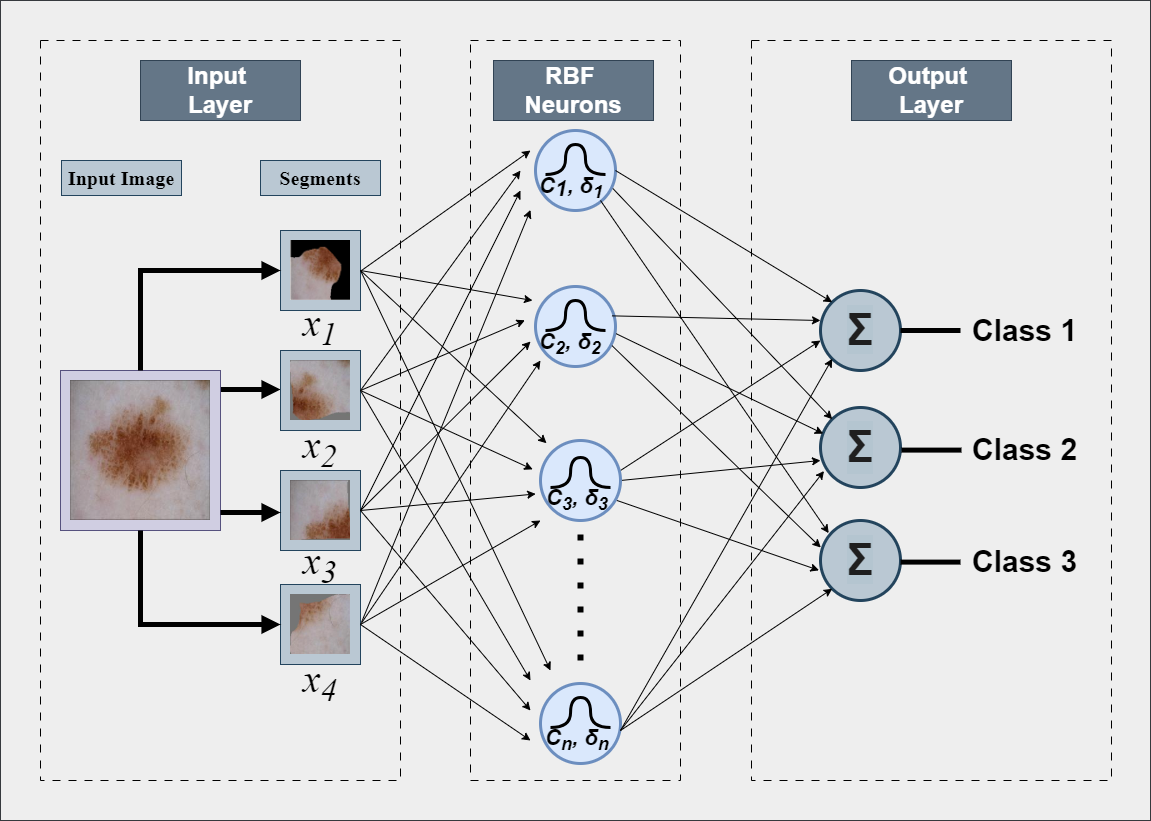}
    \caption{Architecture of proposed Radial basis neural network}
    \label{fig:proposedModel}
\end{figure}

\par
Overall, this model combines the powerful feature extraction capabilities of deep learning with clustering and RBF transformations to create a system that balances high accuracy with interpretability, making it suitable for fine-grained medical image classification tasks. A high-level workflow of the proposed model is shown below in the form of an algorithm provided below in table \ref{tab:algorithm}.

\begin{table}[!htbp]
\centering
\caption{Algorithm for the proposed Radial Basis Function Neural Network (RBF-NN) model for interpretable image classification.}
\begin{tabular}{cl}
\toprule
\multicolumn{2}{l}{\textbf{Inputs:} $X$ = Input Image, $y$ = Class Labels, $k$ = Number of Prototypes, $\sigma$ = Sigma Spread} \\ 
\multicolumn{2}{l}{\textbf{Outputs:} $\hat{y}$ = Predicted Labels, Prototype-Based Explanations} \\ 
\midrule
\textbf{Step} & \textbf{Description} \\ 
\midrule
1 & Segment input image $X$ into $n$ regions: $X = \{X_1, X_2, \dots, X_n\}$. \\ 
2 & Extract embeddings for each segment: $Z_i = f_{\text{CNN}}(X_i)$ for $i = 1, \dots, n$. \\
& Aggregate embeddings: $Z = \{Z_1, Z_2, \dots, Z_n\}$. \\ 
3 & Apply $k$-Medoids clustering on $Z$: $C = \{c_1, c_2, \dots, c_k\}$. \\ 
4 & Compute RBF activation for each embedding $z$: 
$\phi_i(z) = \exp\left(-\frac{\|z - c_i\|^2}{2\sigma^2}\right)$. \\ 
5 & Construct RBF activation vector: $\Phi(z) = [\phi_1(z), \phi_2(z), \dots, \phi_k(z)]^\top$. \\ 
6 & Pass RBF activations to a dense layer: 
$p(y = c \mid z) = \text{softmax}\left(W \cdot \Phi(z) + b\right)$. \\ 
7 & Combine loss functions: \\
   & Cross-Entropy Loss: $\mathcal{L}_{\text{CE}} = -\frac{1}{n} \sum_{j=1}^{n} \sum_{c=1}^{C} y_{j,c} \log\left(p(y = c \mid z_j)\right)$. \\
   & Focal Loss: $\mathcal{L}_{\text{Focal}} = -\frac{1}{n} \sum_{j=1}^{n} \sum_{c=1}^{C} \alpha_c \, y_{j,c} \, (1 - p(y = c \mid z_j))^\gamma \, \log\left(p(y = c \mid z_j)\right)$. \\
   & Total Loss: $\mathcal{L} = \mathcal{L}_{\text{CE}} + \mathcal{L}_{\text{Focal}}$. \\ 
8 & Minimize the loss function: $\mathcal{L} = \min \mathcal{L}(y, \hat{y})$. \\ 
9 & Predict the class label: $\hat{y} = \arg\max_{c} \, p(y = c \mid z)$. \\ 
10 & Trace prediction to most activated prototype: $\text{Prototype Contribution} = \arg\max_{i} \phi_i(z)$. \\ 
\bottomrule
\end{tabular}
\label{tab:algorithm}
\end{table}

This algorithm combines pre-trained feature extraction, clustering for interpretable prototype selection, and RBF-based classification for accurate and explainable predictions. The use of prototypes enables transparency, as each prediction can be traced back to a specific cluster centre, making it well-suited for critical applications like medical diagnosis.

\subsection{Model's training settings}
In this section, we outline the configuration and methodologies employed to train and evaluate the proposed model. The settings are carefully designed to ensure efficient training, robust performance evaluation, and optimal generalization to unseen data. These configurations encompass data preprocessing, model architecture specifics, optimizer settings, loss functions, and evaluation metrics, providing a comprehensive overview of the experimental setup.

\subsubsection{Hardware and Software Configuration}
In this section, we discuss the specifications of the hardware and software we used for this research 

The experiments were conducted using the following hardware and software configurations. The table \ref{tab:system_config} provides the details of the computational environment and libraries used:

\begin{table}[!ht]
\centering
\caption{System Configuration and Libraries}
\begin{tabular}{ll}
\toprule
\textbf{Component}           & \textbf{Details} \\ 
\midrule
\textbf{GPU}                 & NVIDIA T4 (Google Colab) \\ 
\textbf{CPU}                 & Intel Xeon Processor \\ 
\textbf{RAM}                 & 64GB (or as applicable) \\ 
\textbf{Operating System}    & Ubuntu 20.04 LTS \\ 
\textbf{Deep Learning Framework} & PyTorch (version 3.12) \\
\textbf{Additional Libraries} & NumPy, pandas, scikit-learn, Keras, and OpenCV \\ 
\bottomrule
\end{tabular}
\label{tab:system_config}
\end{table}

\subsubsection{Datasets:}
In this study, we utilized the ISIC 2016 and ISIC 2017 skin cancer datasets. The ISIC 2016 dataset, introduced in December 2015, comprises 1,279 dermoscopic images selected from the ISIC Archive. These images were derived after excluding 273 out of the initially selected 1,552 images and are categorized into two classes: benign and malignant. The dataset was randomly split into a training set of 900 images and a test set of 379 images. Images from ISIC 2016 are shown below in figure \ref{fig:ISIC2016_images}.

\begin{figure}[!ht]
    \centering
    \includegraphics[width=\textwidth]{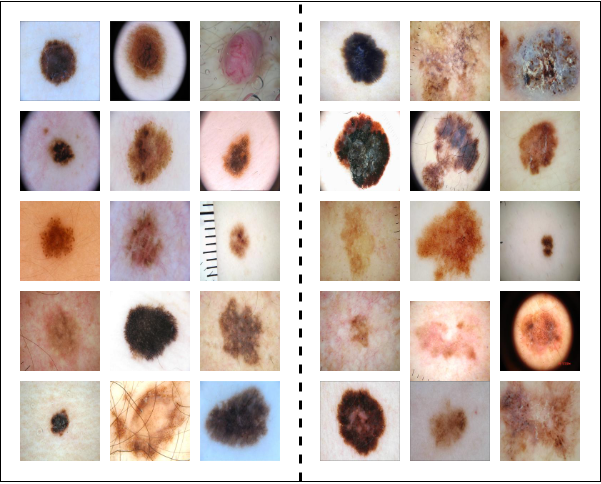}
    \caption{Images from ISIC 2016 of both classes}
    \label{fig:ISIC2016_images}
\end{figure}

\par
The ISIC 2017 dataset contains 2,750 dermoscopic images for skin cancer classification. It is divided into 2,000 training images, 150 validation images, and 600 test images, with ground truth labels and patient metadata available for three classes: melanoma (374 images), nevus (1,372 images), and seborrheic keratosis (254 images). These high-resolution images range in size from 540×722 to 4499×6748 pixels. Images from ISIC 2017 are shown below in figure \ref{fig:ISIC2017_images}.

\begin{figure}[!ht]
    \centering
    \includegraphics[width=\linewidth]{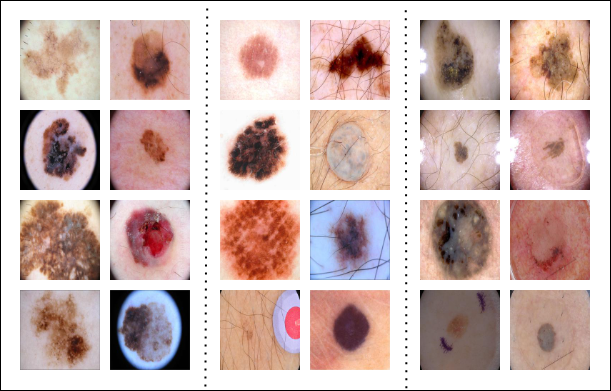}
    \caption{Images from ISIC 2017 of all classes}
    \label{fig:ISIC2017_images}
\end{figure}

\par
Both datasets, despite their relatively small size, were chosen to demonstrate that our proposed model, designed with intrinsically interpretable architecture, achieves superior performance compared to existing approaches.

\subsubsection{Data pre-processing}
All images are resized to a uniform size of 224×224 pixels (as required by the most pre-trained deep learning models). Pixel values are normalized to the range [0, 1], standardized to have a mean of 0 and standard deviation of 1, using the ImageNet dataset mean and standard deviation values for pre-trained models. The superpixel segmentation technique, SLIC, generates distinct regions within each image. In this research, we have created four (04) segments of each image using SLIC with compactness=100 and sigma=1.*3. 

\subsubsection{Concept selection}
In this stage, we select the most salient segments from the newly created dataset of segments. Segments from each class are kept distinct initially. 

\subsubsection{Embedding creation}
Pre-trained models such as ResNet-50, VGG16, EfficientNet, or MedicalNet are used to extract feature embeddings for all the segments of each image from the respective dataset.

\subsubsection{Custom Clustering Layer}
Our proposed RBF neural network uses a custom clustering layering, which selects the embeddings of salient features from each class. Initially, we used 15 prototype embeddings from each class. The selection is based on K-Means clustering, which learns the embeddings from the concepts/segments selected by the users in the previous step. Corresponding images of selected segments from the ISIC 2016 and ISIC 2017 datasets are shown below in figure \ref{fig:ISIC2016_concepts} and figure \ref{fig:ISIC2017_concepts}. 

\begin{figure}[!ht]
    \centering
    \includegraphics[width=\linewidth]{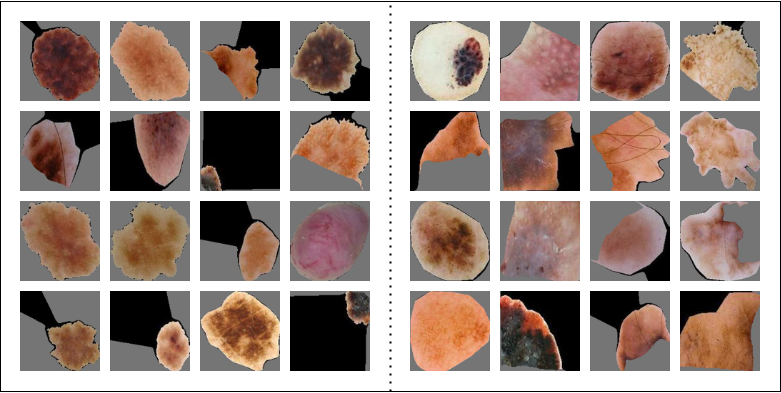}
    \caption{Segments from ISIC 2016 chosen as prototypes by the model from each class}
    \label{fig:ISIC2016_concepts}
\end{figure}

\begin{figure}[!ht]
    \centering
    \includegraphics[width=1\linewidth]{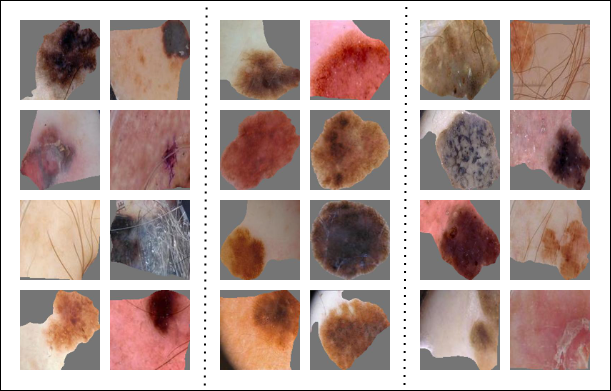}
    \caption{Segments from ISIC 2017 chosen as prototypes by the model from each class}
    \label{fig:ISIC2017_concepts}
\end{figure}

The quality of the clusters is evaluated using silhouette scores and visual inspection of the prototype images. Table \ref{tab:clustering_scores} summarizes the clustering evaluation metrics.

\begin{table}[!ht] 
\centering 
\caption{Clustering quality metrics using K-Medoids for VGG16 and ResNet50 embeddings.} 
\begin{tabular}{lcc} 
\toprule 

\multirow{2}{*}{\textbf{Model Used for embedding creation}} & \multicolumn{2}{c} {\textbf{Silhouette Score }}  \\
         \cmidrule(r){2-3}
 &  \textbf{ISIC 2016} & \textbf{ISIC 2017} \\
 \midrule
 
VGG16 & 0.63 & 0.58 \\ 
 
ResNet50 & 0.72 & 0.66 \\ 

\bottomrule 
\end{tabular} 
\label{tab:clustering_scores} 
\end{table}

As shown in Table \ref{tab:clustering_scores}, ResNet50 produces more coherent clusters with higher silhouette scores compared to VGG16, reflecting its superior feature extraction capabilities.

\subsubsection{Data Augmentation}
In this study, we employed several augmentation techniques to enhance the dataset and improve model performance. Specifically, SMOTE was applied to generate synthetic samples for the minority class, addressing class imbalance by creating new instances that are interpolations of existing minority class samples. Alongside SMOTE, we utilized three distinct image augmentation strategies further to enhance the diversity and robustness of the dataset:

\par
\textbf{Gaussian Noise Augmentation:} This technique adds random Gaussian noise to the computed embeddings of segments, introducing slight variations that help improve model robustness and reduce overfitting by encouraging the model to generalize better.

\begin{equation}
\mathbf{x}_{\text{noisy}} = \mathbf{x} + \mathcal{N}(0, \sigma^2)
\label{eq:gaussian_noise}
\end{equation}

\par
\textbf{Simple Noise Augmentation:} Simple noise is added to the embeddings of image segments by randomly perturbing pixel values, which simulates real-world imperfections in the data, aiding the model in learning to be invariant to minor, random variations.

\begin{equation}
\mathbf{x}_{\text{noisy}}[i,j] = \mathbf{x}[i,j] + \epsilon_{i,j}
\label{eq:simple_noise}
\end{equation}

\par
\textbf{Mixup Augmentation:} Mixup involves combining two or more embeddings by taking a weighted average of their values and corresponding labels. This augmentation technique is effective for improving the model's ability to generalize by presenting it with blended, hybrid instances that might not occur in the original dataset.

\begin{equation}
\mathbf{x}_{\text{mix}} = \lambda \mathbf{x}_i + (1 - \lambda) \mathbf{x}_j
\label{eq:mixup_image}
\end{equation}

\begin{equation}
\mathbf{y}_{\text{mix}} = \lambda \mathbf{y}_i + (1 - \lambda) \mathbf{y}_j
\label{eq:mixup_label}
\end{equation}

By leveraging SMOTE for synthetic sample generation and these noise-based augmentation techniques, we aimed to create a more diverse and balanced dataset, improving the model's performance on skin cancer classification tasks.

\textbf{Loss Function:}
In addition to cross-entropy loss, we utilized categorical focal loss (eq. \ref{eq:loss} to address class imbalance issues commonly encountered in medical datasets. Categorical focal loss extends the traditional focal loss to multi-class classification scenarios. It down-weights the loss assigned to well-classified examples, thereby focusing the training process on hard-to-classify instances. This modification is particularly beneficial in medical imaging, where certain classes may be underrepresented. By emphasizing difficult samples, categorical focal loss enhances the model's ability to learn discriminative features for minority classes, leading to improved sensitivity and overall classification performance. The integration of categorical focal loss with our prototype-based architecture ensures that the model achieves high accuracy and maintains robustness across diverse and imbalanced datasets.

\begin{equation}\label{eq:loss}
\mathcal{L}_{focal} = -\sum_{i=1}^{N} \sum_{c=1}^{C} \alpha_c (1 - \hat{y}_{i,c})^\gamma y_{i,c} \log(\hat{y}_{i,c}),
\end{equation}

\begin{table}[!ht]
\centering
\caption{Training Configuration Overview}
\begin{tabular}{ll}
\toprule
\textbf{Parameter}        & \textbf{Details}     \\ 
\midrule
\textbf{Optimizer}        & Adam with learning rate \(1 \times 10^{-5}\), \(\beta_1 = 0.9\), \(\beta_2 = 0.999\).   \\ 
\textbf{Scheduler}        & Step decay: reduces learning rate every 10 epochs.                                      \\ 
\textbf{Batch Size}       & 32 images (adjusted based on GPU memory).                                               \\ 
\textbf{Epochs}           & Up to 50 with early stopping (patience: 10 epochs).                                     \\ 
\textbf{Validation Split} & Stratified split to ensure balanced class representation.                                \\ 
\textbf{Evaluation Metric} & Accuracy.                                                                              \\ 
\textbf{Mixed Precision}  & Enabled for optimized memory usage and computational efficiency.                        \\ 
\bottomrule
\end{tabular}
\end{table}

\section{Results and discussion}\label{results}
This section presents the findings of the proposed hybrid interpretable deep learning framework for skin cancer diagnosis, followed by an analysis of its implications. The results are evaluated using accuracy. Additionally, we discuss the interpretability aspects of the model, highlighting its ability to generate actionable insights that align with clinical diagnostic practices.

\par
We evaluated the performance of the proposed model on the ISIC 2016 and ISIC 2017 datasets, using accuracy as the primary evaluation metric. In this experiment, the input images from both datasets were segmented, and embeddings for each segment were generated using pre-trained convolutional neural networks (CNNs) such as VGG16 and ResNet50. These embeddings, which capture the most salient features of the segmented images, were subsequently passed to the proposed model for classification and explanation generation. The segmentation process ensures that the model focuses on localized features, which are critical for medical image analysis where subtle patterns and textures play a significant role in diagnosis.

\par
The classification results on the ISIC 2016 and ISIC 2017 datasets are summarized in Table \ref{tab:performance_comparison}, highlighting the model's performance with different CNN architectures and counterpart interpretable models. As shown in Table \ref{tab:performance_comparison}, the use of deeper architectures such as ResNet50 resulted in higher classification accuracy compared to VGG16, emphasizing the importance of advanced feature extraction capabilities. Specifically, the proposed model achieved an accuracy of 83.02\% on ISIC 2016 and 76.15\% on ISIC 2017 when using ResNet50, compared to 79.54\% and 70.09\%, respectively, when using VGG16. These findings validate the effectiveness of the proposed framework in achieving both high accuracy and interpretability for the task of automated skin cancer diagnosis.

\par
When compared to existing interpretable models, the proposed framework demonstrated comparable or superior performance. For instance, ProtoPNet achieved accuracies of 84.50\% on ISIC 2016 and 82.90\% on ISIC 2017, while ProtoPShare achieved accuracies of 92.25\% and 90.10\%, respectively. Although ProtoPShare slightly outperformed the proposed model, it does so at the expense of greater complexity and reduced explainability due to its reliance on shared prototypes. The proposed model, with its integration of RBF networks, provides a balanced approach by ensuring both interpretability and competitive accuracy.

\begin{table}[!ht]
    \centering
     \caption{Performance comparison based accuracy of our proposed framework with related methods on Skin Cancer datasets.}
    \begin{tabular}{lccc}
        \toprule
        \multirow{2}{*}{\textbf{Model Used}} & \multicolumn{2}{c} {\textbf{Dataset}}  \\
         \cmidrule(r){2-3}
        & \textbf{ISIC 2016} & \textbf{ISIC 2017} \\
        \midrule
        ResNet-50 \cite{1} & 92.90\% & 83.00\% \\
        VGG-16 \cite{zunair2020melanoma} & 81.18\% & 74.2\%  \\
        PatchNet-21 \cite{radhakrishnan2017patchnet} & 81.55\% & 75.3\%  \\
        ProtoPNet \cite{ullah2024inherently}  &    84.50     &82.90 \\
        ProtoPShare \cite{ullah2024inherently} &  92.25      & 90.10 \\
        RBF-NN (VGG-16)    & 79.54\%            & 70.09\%          \\ 
        RBF-NN (ResNet-50) & 83.02\%  & 76.15\%  \\    
        \bottomrule
    \end{tabular}
   
    \label{tab:performance_comparison}
\end{table}

\par
The results also underscore the importance of leveraging advanced feature extraction capabilities of deeper CNN architectures. The improvement in performance when using ResNet50 can be attributed to its residual connections, which mitigate the vanishing gradient problem and enable the capture of more complex features. This capability is particularly beneficial in medical imaging tasks, where subtle variations in texture and patterns play a critical role in diagnosis. Despite the superior performance of ResNet50, VGG16 demonstrated competitive results, suggesting that shallower architectures may still be viable for specific applications, especially where computational resources are limited.

\par
\begin{figure}[!ht]
\centering
\includegraphics[width=\linewidth]{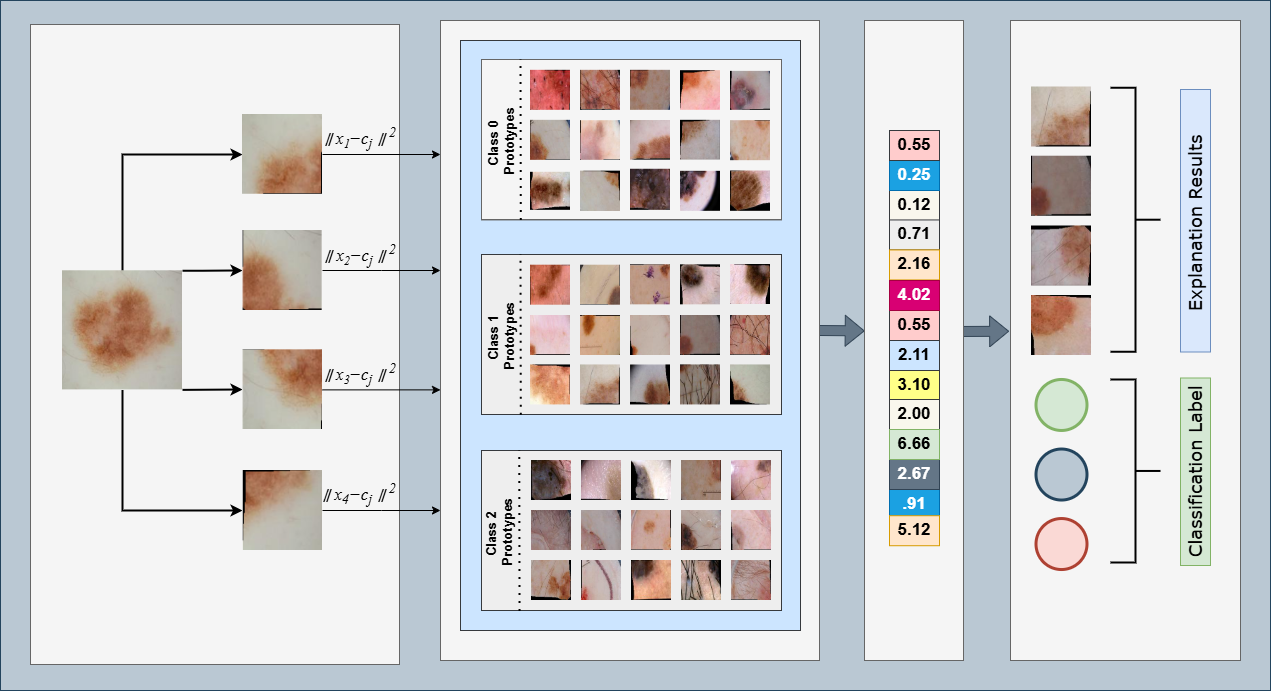}
\caption{Visualization of results and explanations of the proposed model}
\label{fig:final_results}
\end{figure}

The proposed framework's key strength lies in its interpretability. By integrating Radial Basis Function (RBF) Networks with CNNs, the model provides localized explanations for its predictions, addressing the "black box" nature of traditional deep learning models. The use of prototype-based clustering ensures that the model's decision-making process is aligned with clinically relevant features, thereby enhancing trustworthiness and facilitating adoption in healthcare settings. The ability to trace predictions back to specific prototypes provides actionable insights that can support clinical decision-making and improve diagnostic accuracy. The detailed demonstration of the proposed model is shown in Figure \ref{fig:final_results} provided below.

\par
The inclusion of active learning and K-Medoids clustering further enhances the interpretability of the model by selecting salient features that represent each class. This approach improves classification performance and enables the generation of explanations that are faithful to the data and understandable to end users. However, the reliance on predefined clustering methods introduces a potential limitation, as the quality of the clusters may vary depending on the dataset and parameter settings.

\par
While the proposed model demonstrates strong performance and interpretability, several limitations must be acknowledged. First, the relatively small size and class imbalance problem of the ISIC datasets may limit the generalizability of the findings. Although data augmentation techniques and active learning were employed to mitigate this issue, further validation on larger and more diverse datasets is necessary. Second, the computational complexity of the framework, particularly the clustering and RBF layers, may pose challenges for deployment in resource-constrained environments. Future work could focus on optimizing the model architecture to reduce computational overhead without compromising performance.

\par
The interpretability provided by the model is another crucial factor, particularly for high-stakes applications like healthcare. By offering localized explanations through the RBF layer, clinicians can better understand the rationale behind a prediction. This aligns with the growing demand for explainable AI in medical diagnostics, where trust and transparency are paramount. Moreover, the active learning component enables domain experts to guide the model by selecting relevant features, which ensures that the explanations are grounded in clinical knowledge. This collaborative approach between AI systems and medical practitioners could significantly enhance diagnostic workflows.

\par
Another aspect worth noting is the scalability of the proposed framework. Although the current implementation focuses on skin cancer diagnosis, the model's architecture can be adapted to other medical imaging tasks with similar interpretability requirements. For example, the framework could be extended to applications in histopathology or radiology, where the need for transparent and accurate diagnostic tools is equally critical. This flexibility underscores the broader applicability of the proposed approach beyond dermatology.

\par
The discussion also highlights the potential for integrating additional interpretability mechanisms, such as concept-based explanations or saliency maps, to complement the prototype-based approach. These enhancements could provide a multi-faceted understanding of the model's predictions, further increasing its utility in clinical settings. Additionally, incorporating feedback loops for continuous learning could help the model adapt to evolving diagnostic criteria or new data, ensuring its long-term relevance and effectiveness.

\par
In summary, the results of this study highlight the potential of the proposed hybrid framework to balance accuracy and interpretability in automated skin cancer diagnosis. The findings emphasize the importance of integrating explainable AI techniques with deep learning to develop reliable and clinically valuable diagnostic tools. Future research should focus on addressing the identified limitations, exploring the application of this framework to other medical imaging tasks, and investigating ways to enhance its scalability and interpretability.

\section{Conclusion}\label{conclusion}
This study proposed a hybrid interpretable deep learning framework combining the feature extraction strengths of CNNs with the transparency of Radial Basis Function (RBF) Networks for skin cancer diagnosis. Addressing the need for explainable AI in healthcare, the framework tackles the "black box" nature of traditional models by integrating segmentation-based embeddings, prototype clustering, and active learning to achieve high accuracy while ensuring interpretability.

\par Traditional diagnostic methods, reliant on clinical expertise and histopathology, are often subjective and labor-intensive. While existing deep learning models excel in classification, they lack the transparency needed for clinical use. Our framework embeds explainable AI techniques into the architecture, aligning with clinical workflows and enhancing usability.

\par Tests on ISIC 2016 and ISIC 2017 datasets showed ResNet50 achieved the highest accuracy (83.02\% and 76.15\%, respectively), emphasizing the role of deep architectures in detecting subtle diagnostic patterns. The prototype-based decision-making provided localized, interpretable insights, boosting trust and facilitating clinical adoption. Active learning and K-Medoids clustering enhanced interpretability by selecting salient features, while segmentation and prototype clustering focused the model on clinically relevant data, improving performance and transparency. These features make the framework suitable for resource-limited settings.

\par However, limitations include the small ISIC datasets and the computational complexity of clustering and RBF layers, which may hinder deployment in constrained environments. Future work should optimize scalability, validate results on larger datasets, and explore alternative clustering techniques to enhance robustness and clinical applicability.

\par In summary, this framework bridges the gap between predictive accuracy and clinical applicability, offering a step forward in explainable AI for skin cancer diagnosis. With further refinement, it could drive broader adoption of AI diagnostic systems, improving patient care and clinical efficiency.

\bibliographystyle{unsrt}  
\bibliography{references}  

\begin{thebibliography}{10}

\bibitem{1}
Xiaoxuan Liu and et~al.
\newblock A comparison of deep learning performance against health-care professionals in detecting diseases from medical imaging: a systematic review and meta-analysis.
\newblock {\em The Lancet Digital Health}, 1(6):e271--e297, 2019.

\bibitem{2}
Sara Hooker and et~al.
\newblock A benchmark for interpretability methods in deep neural networks.
\newblock In {\em Advances in Neural Information Processing Systems}, volume~32, 2019.

\bibitem{3}
RC~Maron, JG~Schlager, S~Haggenmüller, and et~al.
\newblock A benchmark for neural network robustness in skin cancer classification.
\newblock {\em European Journal of Cancer}, 155:191--199, 2021.

\bibitem{4}
RC~Maron, S~Haggenmüller, C~von Kalle, and et~al.
\newblock Robustness of convolutional neural networks in recognition of pigmented skin lesions.
\newblock {\em European Journal of Cancer}, 145:81--91, 2021.

\bibitem{5}
{European Commission}.
\newblock White paper on artificial intelligence - a european approach to excellence and trust, 2020.
\newblock Accessed: 28 December 2021.

\bibitem{6}
Amy McGovern and et~al.
\newblock Making the black box more transparent: Understanding the physical implications of machine learning.
\newblock {\em Bulletin of the American Meteorological Society}, 100(11):2175--2199, 2019.

\bibitem{7}
Benjamin~A. Toms, Elizabeth~A. Barnes, and Imme Ebert-Uphoff.
\newblock Physically interpretable neural networks for the geosciences: Applications to earth system variability.
\newblock {\em Journal of Advances in Modeling Earth Systems}, 12(9):e2019MS002002, 2020.

\bibitem{8}
Ao~Li and et~al.
\newblock Attention-based interpretable neural network for building cooling load prediction.
\newblock {\em Applied Energy}, 299:117238, 2021.

\bibitem{9}
Botao An and et~al.
\newblock Interpretable neural network via algorithm unrolling for mechanical fault diagnosis.
\newblock {\em IEEE Transactions on Instrumentation and Measurement}, 71:1--11, 2022.

\bibitem{10}
Christoph Molnar and et~al.
\newblock General pitfalls of model-agnostic interpretation methods for machine learning models.
\newblock In {\em International Workshop on Extending Explainable AI Beyond Deep Models and Classifiers}. Springer International Publishing, 2020.

\bibitem{11}
Grégoire Montavon, Wojciech Samek, and Klaus-Robert Müller.
\newblock Methods for interpreting and understanding deep neural networks.
\newblock {\em Digital Signal Processing}, 73:1--15, 2018.

\bibitem{gutman2016skin}
David Gutman, Noel~CF Codella, Emre Celebi, Brian Helba, Michael Marchetti, Nabin Mishra, and Allan Halpern.
\newblock Skin lesion analysis toward melanoma detection: A challenge at the international symposium on biomedical imaging (isbi) 2016, hosted by the international skin imaging collaboration (isic).
\newblock {\em arXiv preprint arXiv:1605.01397}, 2016.

\bibitem{codella2018skin}
Noel~CF Codella, David Gutman, M~Emre Celebi, Brian Helba, Michael~A Marchetti, Stephen~W Dusza, Aadi Kalloo, Konstantinos Liopyris, Nabin Mishra, Harald Kittler, et~al.
\newblock Skin lesion analysis toward melanoma detection: A challenge at the 2017 international symposium on biomedical imaging (isbi), hosted by the international skin imaging collaboration (isic).
\newblock In {\em 2018 IEEE 15th international symposium on biomedical imaging (ISBI 2018)}, pages 168--172. IEEE, 2018.

\bibitem{bolhasani2021deep}
Hamidreza Bolhasani, Maryam Mohseni, and Amir~Masoud Rahmani.
\newblock Deep learning applications for iot in health care: A systematic review.
\newblock {\em Informatics in Medicine Unlocked}, 23:100550, 2021.

\bibitem{esteva2019guide}
Andre Esteva, Alexandre Robicquet, Bharath Ramsundar, Volodymyr Kuleshov, Mark DePristo, Katherine Chou, Claire Cui, Greg Corrado, Sebastian Thrun, and Jeff Dean.
\newblock A guide to deep learning in healthcare.
\newblock {\em Nature medicine}, 25(1):24--29, 2019.

\bibitem{pandey2019recent}
Saroj~Kumar Pandey and Rekh~Ram Janghel.
\newblock Recent deep learning techniques, challenges and its applications for medical healthcare system: a review.
\newblock {\em Neural Processing Letters}, 50(2):1907--1935, 2019.

\bibitem{kaul2022deep}
Deeksha Kaul, Harika Raju, and BK~Tripathy.
\newblock Deep learning in healthcare.
\newblock {\em Deep Learning in Data Analytics: Recent Techniques, Practices and Applications}, pages 97--115, 2022.

\bibitem{chakraborty2024machine}
Chiranjib Chakraborty, Manojit Bhattacharya, Soumen Pal, and Sang-Soo Lee.
\newblock From machine learning to deep learning: Advances of the recent data-driven paradigm shift in medicine and healthcare.
\newblock {\em Current Research in Biotechnology}, 7:100164, 2024.

\bibitem{maqsood2023multiclass}
Sarmad Maqsood and Robertas Dama{\v{s}}evi{\v{c}}ius.
\newblock Multiclass skin lesion localization and classification using deep learning based features fusion and selection framework for smart healthcare.
\newblock {\em Neural networks}, 160:238--258, 2023.

\bibitem{shafqat2023leveraging}
Sarah Shafqat, Maryyam Fayyaz, Hasan~Ali Khattak, Muhammad Bilal, Shahid Khan, Osama Ishtiaq, Almas Abbasi, Farzana Shafqat, Waleed~S Alnumay, and Pushpita Chatterjee.
\newblock Leveraging deep learning for designing healthcare analytics heuristic for diagnostics.
\newblock {\em Neural processing letters}, pages 1--27, 2023.

\bibitem{zhou2016learning}
Bolei Zhou, Aditya Khosla, Agata Lapedriza, Aude Oliva, and Antonio Torralba.
\newblock Learning deep features for discriminative localization.
\newblock In {\em Proceedings of the IEEE conference on computer vision and pattern recognition}, pages 2921--2929, 2016.

\bibitem{chattopadhay2018grad}
Aditya Chattopadhay, Anirban Sarkar, Prantik Howlader, and Vineeth~N Balasubramanian.
\newblock Grad-cam++: Generalized gradient-based visual explanations for deep convolutional networks.
\newblock In {\em 2018 IEEE winter conference on applications of computer vision (WACV)}, pages 839--847. IEEE, 2018.

\bibitem{zhang2024opti}
Hanwei Zhang, Felipe Torres, Ronan Sicre, Yannis Avrithis, and Stephane Ayache.
\newblock Opti-cam: Optimizing saliency maps for interpretability.
\newblock {\em Computer Vision and Image Understanding}, 248:104101, 2024.

\bibitem{he2016deep}
Kaiming He, Xiangyu Zhang, Shaoqing Ren, and Jian Sun.
\newblock Deep residual learning for image recognition.
\newblock In {\em Proceedings of the IEEE conference on computer vision and pattern recognition}, pages 770--778, 2016.

\bibitem{zunair2020melanoma}
Hasib Zunair and A~Ben Hamza.
\newblock Melanoma detection using adversarial training and deep transfer learning.
\newblock {\em Physics in Medicine \& Biology}, 65(13):135005, 2020.

\bibitem{simonyan2014very}
Karen Simonyan.
\newblock Very deep convolutional networks for large-scale image recognition.
\newblock {\em arXiv preprint arXiv:1409.1556}, 2014.

\bibitem{achtibat2023attribution}
Reduan Achtibat, Maximilian Dreyer, Ilona Eisenbraun, Sebastian Bosse, Thomas Wiegand, Wojciech Samek, and Sebastian Lapuschkin.
\newblock From attribution maps to human-understandable explanations through concept relevance propagation.
\newblock {\em Nature Machine Intelligence}, 5(9):1006--1019, 2023.

\bibitem{r-5}
Satyapriya Krishna, Tessa Han, Alex Gu, Javin Pombra, Shahin Jabbari, Steven Wu, and Himabindu Lakkaraju.
\newblock The disagreement problem in explainable machine learning: A practitioner’s perspective.
\newblock {\em arXiv preprint arXiv:2202.01602}, 2022.

\bibitem{r-6}
Vinitra Swamy, Bahar Radmehr, Natasa Krco, Mirko Marras, and Tanja Käser.
\newblock Evaluating the explainers: Black-box explainable machine learning for student success prediction in moocs.
\newblock In {\em International Conference on Educational Data Mining}, 2022.

\bibitem{r-6-18}
Vinitra Swamy and et~al.
\newblock Interpretcc: Intrinsic user-centric interpretability through global mixture of experts, 2024.

\bibitem{r-7}
Cynthia Rudin.
\newblock Stop explaining black box machine learning models for high stakes decisions and use interpretable models instead.
\newblock {\em Nature Machine Intelligence}, 1:206--215, 2019.

\bibitem{r-8}
Zachary~C. Lipton.
\newblock The mythos of model interpretability.
\newblock {\em Queue}, 16(3):31--57, 2018.

\bibitem{quinlan1986induction}
J.~Ross Quinlan.
\newblock Induction of decision trees.
\newblock {\em Machine learning}, 1:81--106, 1986.

\bibitem{rokach2005decision}
Lior Rokach and Oded Maimon.
\newblock Decision trees.
\newblock {\em Data mining and knowledge discovery handbook}, pages 165--192, 2005.

\bibitem{khashei2012novel}
Mehdi Khashei, Ali~Zeinal Hamadani, and Mehdi Bijari.
\newblock A novel hybrid classification model of artificial neural networks and multiple linear regression models.
\newblock {\em Expert Systems with Applications}, 39(3):2606--2620, 2012.

\bibitem{van2017fifty}
Bram van Ginneken.
\newblock Fifty years of computer analysis in chest imaging: rule-based, machine learning, deep learning.
\newblock {\em Radiological physics and technology}, 10:23--32, 2017.

\bibitem{glaab2012using}
Enrico Glaab, Jaume Bacardit, Jonathan~M Garibaldi, and Natalio Krasnogor.
\newblock Using rule-based machine learning for candidate disease gene prioritization and sample classification of cancer gene expression data.
\newblock {\em PloS one}, 7(7):e39932, 2012.

\bibitem{buhmann2000radial}
Martin~Dietrich Buhmann.
\newblock Radial basis functions.
\newblock {\em Acta numerica}, 9:1--38, 2000.

\bibitem{majdisova2017radial}
Zuzana Majdisova and Vaclav Skala.
\newblock Radial basis function approximations: comparison and applications.
\newblock {\em Applied Mathematical Modelling}, 51:728--743, 2017.

\bibitem{chen2019looks}
Chaofan Chen, Oscar Li, Daniel Tao, Alina Barnett, Cynthia Rudin, and Jonathan~K Su.
\newblock This looks like that: deep learning for interpretable image recognition.
\newblock {\em Advances in neural information processing systems}, 32, 2019.

\bibitem{rymarczyk2021protopshare}
Dawid Rymarczyk, {\L}ukasz Struski, Jacek Tabor, and Bartosz Zieli{\'n}ski.
\newblock Protopshare: Prototypical parts sharing for similarity discovery in interpretable image classification.
\newblock In {\em Proceedings of the 27th ACM SIGKDD Conference on Knowledge Discovery \& Data Mining}, pages 1420--1430, 2021.

\bibitem{ma2023looks}
Chiyu Ma, Brandon Zhao, Chaofan Chen, and Cynthia Rudin.
\newblock This looks like those: Illuminating prototypical concepts using multiple visualizations.
\newblock {\em Advances in Neural Information Processing Systems}, 36:39212--39235, 2023.

\bibitem{zia2022softnet}
Tehseen Zia, Nauman Bashir, Mirza~Ahsan Ullah, and Shakeeb Murtaza.
\newblock Softnet: A concept-controlled deep learning architecture for interpretable image classification.
\newblock {\em Knowledge-Based Systems}, 240:108066, 2022.

\bibitem{ullah2024inherently}
Mirza~Ahsan Ullah, Tehseen Zia, Jungeun Kim, and Seifedine Kadry.
\newblock An inherently interpretable deep learning model for local explanations using visual concepts.
\newblock {\em Plos one}, 19(10):e0311879, 2024.

\bibitem{radhakrishnan2017patchnet}
Adityanarayanan Radhakrishnan, Charles Durham, Ali Soylemezoglu, and Caroline Uhler.
\newblock Patchnet: interpretable neural networks for image classification.
\newblock {\em arXiv preprint arXiv:1705.08078}, 2017.

\end{thebibliography}


\end{document}